\documentclass[letterpaper, 10 pt, journal, twoside]{IEEEtran}
\IEEEoverridecommandlockouts                              

\usepackage{graphics} 
\usepackage{epsfig} 
\usepackage{mathptmx} 
\usepackage{times} 
\usepackage{amsmath} 
\usepackage{amssymb}  

\usepackage{subcaption} 
\usepackage{listings}
\usepackage{algorithm}
\usepackage{algorithmic}

\usepackage{siunitx}
\usepackage[normalem]{ulem} 

\usepackage{xcolor}
\usepackage{soul}

\usepackage{hyperref}

\usepackage[font=small, labelfont=bf, belowskip=-2pt]{caption} 

\usepackage[
backend=biber,  
style=ieee,
sorting=none
]{biblatex}

\title{
  Fast Obstacle Avoidance \\
  Based on Real-Time Sensing
}

\author{Lukas Huber$^{1}$
  $\cdot$ Jean-Jacques Slotine $^2$ $\cdot$ Aude Billard $^1$ \vspace{-2.3ex}
\thanks{*This work was funded in part by the EU ERC grant SAHR.}
\thanks{$^{1}$ LASA Laboratory, Swiss Federal School of Technology in Lausanne - EPFL, Switzerland. \tt $\{$lukas.huber;aude.billard$\}$@epfl.ch }
\thanks{$^{3}$ Nonlinear Systems Laboratory,  Massachusetts Institute of Technology, USA. \tt jjs@mit.edu}   
}

\begin{document}
\newcommand{\vect}[1]{\mathbf{#1}}
\newcommand{\matr}[1]{\mathbf{#1}}

\newcommand{\dotprod}[2]{\langle {#1}, \, {#2} \rangle}
\newcommand{\normdotprod}[2]{\frac{\langle #1, \, #2 \rangle}{\| #1 \| \, \| #2 \|}}

\newcommand{\editcolor}[1]{\color{blue}{#1} \color{black}}
\newcommand{\tempcolor}[1]{\color{violet}{#1} \color{black}}

\maketitle
\thispagestyle{empty}  

\begin{abstract}
Humans are remarkable at navigating and moving through dynamic and complex spaces, such as crowded streets. For robots to do the same, it is crucial that they are endowed with highly reactive obstacle avoidance robust to partial and poor sensing. 
We address the issue of enabling obstacle avoidance based on sparse and asynchronous perception. 
The proposed control scheme combines a high-level input command provided by either a planner or a human operator with fast reactive obstacle avoidance. 
The sampling-based sensor data can be combined with an analytical reconstruction of the obstacles for real-time collision avoidance. 
We can ensure that the agent does not get stuck when a feasible path exists between obstacles. 
The algorithm was evaluated experimentally on static laser data from cluttered, indoor office environments. Additionally, it was used in a shared control mode in a dynamic and complex outdoor environment in the center of Lausanne.
    The proposed control scheme successfully avoided collisions in both scenarios.
  During the experiments, the controller on the onboard computer took 1 millisecond to evaluate over 30000 data points. 
\end{abstract}

\section{Introduction} \label{sec:introduction}
Local avoidance algorithms have to take into account new sensory information in real-time and adapt the global path to ensure safe navigation in the environment. 

One approach to elevate distance sensor data is the Vector Field Histograms (VFH) method, which has been applied on mobile robots to navigate autonomously \cite{borenstein1991vector}. VFH has the advantage of taking measurement uncertainties into account. Furthermore, the short evaluation time of VFH makes it applicable to shared control on wheelchairs \cite{levine1999navchair, li2011dynamic}. However, VFH has the shortcoming of the control velocity being zero in narrow pathways, for example, in front of (narrow) doorways. 

An alternative set of approaches are sampling-based methos. They have been used for fast self-collision avoidance of robots in joint space \cite{mirrazavi2018unified, koptev2021real}. The collision-free environment is taught using the sampled data. As the models require training before execution, the methods cannot adapt to fast-changing environments. 

Conversely, the Velocity Obstacle (VO) approach is designed for dynamic environments. VO takes into account the motion of the obstacles and their potential future positions to calculate a collision-free motion \cite{fiorini1998motion, wilkie2009generalized, kufoalor2018proactive}. VO has been successfully used in shared control on wheelchairs \cite{prassler2001robotics}. However, VO is prone to local minima, which predominantly becomes apparent in static, cluttered environments. 

Recently, real-time optimization algorithms have become feasible for implementing onboard robots thanks to the improved computational performance of hardware, for example, model predictive control \cite{bardaro2018mpc} or quadratic programming \cite{gonon2021reactive}. However, most optimization problems cannot guarantee convergence to a feasible solution. 

Recent work on Control Barrier Functions (CBFs) can ensure a feasible solution by softening some constraints \cite{ames2014control} or using slack variables \cite{reis2020control}. Nevertheless, these CBFs require a Lyapunov candidate function to reach the goal position and cannot be applied to dynamic environments. 

Conversely, artificial potential fields create a closed-form, repulsive field to avoid collision with obstacles \cite{khatib1986real}. They can be elevated with navigation functions to ensure convergence to the desired goal around concave obstacles \cite{rimon1992exact, zhang2019distributed}. 
Recent work extends the navigation function to discovering obstacles' positions at runtime, but the approach relies on prior knowledge about the obstacles' shapes \cite{loizou2022mobile}.
However, navigation functions are difficult to tune around dynamic obstacles. Moreover, no approach exists that combines the method with high-level control inputs, as is the case for shared control. 

Similarly, \cite{feder1997real} creates a dynamical system based algorithm inspired by harmonic potential fields, which can ensure collision-free motion in dynamic environments. 
This has been extended to ensure convergence around concave obstacles by taking a (linear) velocity towards a goal and guiding it around all obstacles \cite{khansari2012dynamical, huber2019avoidance, huber2021avoiding}.
However, the approach requires the analytical description of the obstacles, and interpreting the sensor data onboard a robot often introduces a significant delay \cite{huber2021avoiding}.

\subsection{Contribution}
We propose a fast, dynamical system-based control algorithm based on \cite{huber2019avoidance, huber2021avoiding} with the following new contributions:
\begin{itemize}
\item \textbf{Low computational complexity} ensures fast evaluation in a cluttered environment, such as crowds (see Sec.~\ref{sec:non_circular}).

\item \textbf{Obstacle avoidance using sampled data} makes the method directly applicable to Lidar data (see Sec.~\ref{sec:single_modulation}).

\item \textbf{Unifying disparate obstacle descriptions} such as sample-based sensor data and (delayed) analytic reconstruction of obstacles is enabled through asynchronous evaluation and dynamic weighting (see Sec.~\ref{sec:mixed_environments}).
\end{itemize}

\section{Preliminaries}
The state of the robotics system $\xi \in \mathbb{R}^d$ refers to the position of a $d$-dimensional state throughout this work. 
The evolution of the position is governed by a dynamical system 
\begin{equation}
   \dot \xi = \vect f(t, \xi)  \quad \text{with} \quad  \vect f(t, \xi) : (\mathbb{R}_+, \mathbb{R}^d) \rightarrow  \mathbb{R}^d
\end{equation}
We further introduce a linear (autonomous) dynamical system which is asymptotically stable at the attractor position $\xi^a$: 
\begin{equation}
\vect f^l(\xi) = -(\xi - \xi^a) 
\qquad \text{with} \quad
\xi^ \in \mathbb{R}^d \label{eq:linear_dynamics}
\end{equation} 

\subsection{Obstacles and Gamma Function} \label{sec:gamma_function}
As in \cite{khansari2012dynamical}, a continuous distance function $\Gamma(\xi): \mathbb{R}^d \setminus \mathcal{X}^i \mapsto \mathbb{R}_{\geq 1}$ is defined around each obstacle, and divides the space into three regions:
\begin{align}
  &\text{Exterior points:}&  \;\; & \mathcal{X}^e = \{\xi \in \mathbb{R}^d: \Gamma(\xi) > 1 \} \nonumber \\ 
  &\text{Margin exterior points:}  & \;\; & \mathcal{X}^g = \{\xi - (R + D^{\mathrm{gap}}) \vect r(\xi) \in \mathcal{X}^e \} \nonumber \\
  &\text{Boundary points:}&  \;\; & \mathcal{X}^b = \{\xi \in \mathbb{R}^d: \Gamma(\xi) = 1 \} \label{eq:gamma_function} \\
  &\text{Interior points:}&  \;\; & \mathcal{X}^i = \{ \xi \in \mathbb{R}^d \setminus (  \mathcal{X}^e \cup \mathcal{X}^b ) \} \nonumber
\end{align}
where $D^{\mathrm{gap}} \in \mathbb{R}_{>0}$ is the margin distance, and $R \in \mathbb{R}_{>0}$ the robot radius.
By construction $\Gamma(\cdot)$ increases monotonically ($C^1$-smoothness) with increasing distance from the center $\xi^c \in \mathbb{R}^d$.

The obstacle avoidance algorithm \cite{huber2019avoidance} is defined for \textit{star-shapes}, i.e., a shape where there exists a reference point $\xi^r \in \mathbb{R}^d$ within the obstacle, such that any surface point can be seen:
\begin{equation}
\exists \, \xi^r \in \mathcal{X}^i
\quad : \;\;
 \dotprod{\vect r(\xi)}{\vect n(\xi)} > 0 \;\;\;\; \forall \xi \in \mathcal{X}^e \label{eq:star_shape}
\end{equation}
where $\vect r(\xi) = (\xi - \xi^r)/\| \xi - \xi^r\|$ with $\vect r(\xi) \in \mathbb{R}^d, \; \| \vect r(\xi) \| = 1$ is denoted as the normalized reference direction.
An environment made up of \textit{star-shaped} obstacles is referred to as a \textit{star world}.

\subsection{Obstacle Avoidance through Modulation}
Assuming the agents nominal velocity command of $\vect v^N \in \mathbb{R}^d$, which restuls from a user's input or a high-level planner evaluated at position $\xi$. 
As in \cite{huber2019avoidance}, the collision free velocity around a single obstacle is obtained by applying a dynamic modulation matrix $\matr M\in \mathbb{R}^{d \times d}$:
\begin{equation}
  \dot{\xi} = \matr{M}(\xi, \vect v^N) \vect v^N
   \; \text{,} \quad
  \matr{M} ( \xi, \vect v^N) = \matr{E}(\xi) \matr D(\xi, \vect v^N) \matr{E}(\xi)^{-1}
  \label{eq:modulated_ds}
\end{equation}
The basis matrix is defined as:
\begin{equation}
  \matr {E} (\xi) =
  \left[ {\vect r }(\xi) \;\; \vect e_1(\xi) \;\; .. \;\; \vect{e}_{d-1}(\xi) \right]
   \label{eq:basisMatr}
\end{equation}
where $\vect e_{(\cdot)}$ form an orthonormal basis of the tangent plane.\footnote{For a spherical obstacle, decomposition matrix $\matr E$ is orthonormal, hence $\matr{E}^{-1} = \matr{E}^{T}$.} 
The diagonal eigenvalue matrix $\matr D(\xi)$ defines the stretching in each direction.
\section{Fast Obstacle Avoidance} \label{sec:non_circular}
When navigating in cluttered environments, such as dense crowds, the algorithm described \cite{huber2021avoiding}, requires applying the modulation for all obstacles at each time step. This results in a high computational cost which increases linearly with the number of obstacles.
 We propose a computationally cheaper algorithm by creating a single \textit{virtual} obstacle that encapsulates all obstacles. The modulation is then only applied to these obstacles while ensuring collision avoidance for all obstacles. The virtual obstacle has a single modulation matrix $\matr M(\xi)$ with corresponding decomposition matrix $\matr E(\xi)$ and diagonal stretching matrix $\matr D(\xi)$

\subsection{Single Modulation Obstacle Avoidance} \label{sec:single_modulation_avoidance}
To evaluate the decomposition matrix $\matr E(\xi)$ as defined in (\ref{eq:basisMatr}), we use the normalized reference direction $\vect{r}^o(\xi) = \hat{\vect r}^o(\xi) / \| \hat{\vect r}^o(\xi) \| $, and the tangent directions $\vect e_i^o(\xi)$ form the orthonormal basis to $\vect{n}^o(\xi)$. The vectors are further described in the following subsections.

\subsubsection{Averaged Reference Direction}
The averaged reference direction $\hat{\vect{r}}^o(\xi)$ is evaluated as the weighted sum:
\begin{equation}
    \hat{\vect{r}}^o(\xi) = \sum_i w_i^o(\xi) \vect r_i^o(\xi) 
    \quad i = 1 .. N^{\mathrm{obs}}
    \label{eq:reference_summing}
\end{equation}
where the reference directions $\vect{r}^o_i(\xi)$ are evaluated for each obstacle as described in (\ref{eq:star_shape}).
Additionally, the influence weights $w_i^o(\xi)$ are given by 
\begin{multline}
 w^o_i(\xi) =
 \begin{cases}
 \hat w^o_i(\xi) / \hat w^{\mathrm{sum}} \quad & \text{if}  \;\; \hat w^{\mathrm{sum}} > 1 \\
 {w}^o_i(\xi)  & \text{otherwise}
 \end{cases}
 \, , \;
 w^{\mathrm{sum}} = \sum_i \hat w_i^o(\xi)
  \label{eq:weight_sum_obstacle}
\end{multline}
which is a function of the distance Gamma, see (\ref{eq:gamma_function}), and evaluated for each obstacle $i$ as follows:
\begin{equation}
    \hat{w}^o_i(\xi) = \left( \frac{D^{\mathrm{scal}}}{D^o_i(\xi)} \right)^s
    \qquad \text{with} \quad
    D_i^o (\xi) = \Gamma_i(\xi) - 1 
\end{equation}
where $s \in \mathbb{R}_{>0}$ is the scaling-potential and $D^{\mathrm{scal}} \in \mathbb{R}_{>0}$. We choose $s = 2$ and $D^{\mathrm{scal}}= 1$.



\subsubsection{Summing of the Normal Direction} \label{sec:normal_offset}

First, the normal offset is evaluated for all obstacles
\begin{equation}
  {\vect n}^{\Delta}(\xi) = \sum^{N^{\mathrm{obs}}}_{i=1} w_i^o(\xi) \left( \vect{n}_i(\xi) - \vect{r}^o_i(\xi) \right)
\end{equation}
where $N^{\mathrm{obs}} \in \mathbb{N}_{\geq 0}$ is the number of obstacles.
Finally, the weighted normal vector is obtained as:
\begin{gather}
  {\vect{n}}(\xi) = \hat{\vect{n}}(\xi) / \| \hat{\vect{n}}(\xi)\|
  \quad \text{with} \;\;
  \hat{\vect{n}}(\xi) = c^n \vect{r}(\xi) +  {\vect n}^{\Delta}(\xi) \label{eq:normal_vector} \\
    c^n = \begin{cases}
    1 & \text{if} \;\;  p^{nr} < \frac{\sqrt{2}}{2} \\
    \sqrt{2} p^{nr}& \text{otherwise}
  \end{cases}
  \;\; \text{with} \; p^{nr} = (-1)  
  \frac{\dotprod{\vect{r}^o(\xi)}{ \vect n^{\Delta}(\xi) } }{\| \vect n^{\Delta}(\xi) \|} \nonumber
\end{gather}
The scaling factor $c^n \in [1, \sqrt{2} ]$ ensures that the decomposition matrix $\matr E(\xi)$ is invertible, see Appendix~\ref{sec:proof_theorem1} for further explanation. 

\subsection{Eigenvalue Matrix}
\begin{figure}[t]
\centering
\includegraphics[width=1.0\columnwidth]{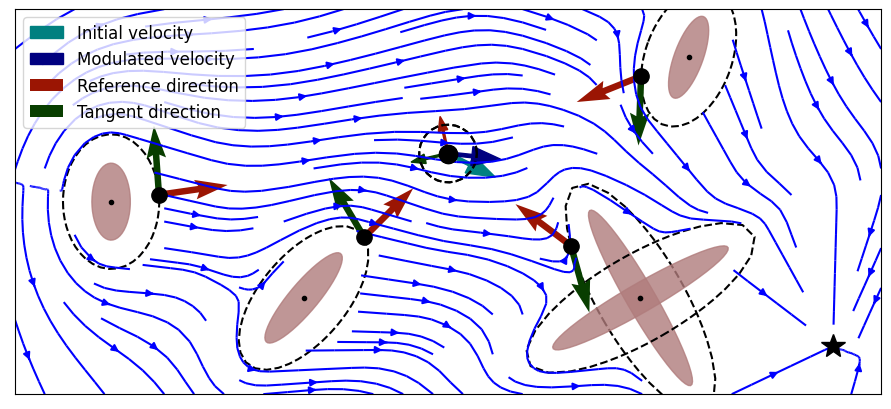}
\caption{In the surrounding with analytical obstacle descriptions (brown), a \textit{virtual} obstacle is constructed with corresponding reference and tangent direction. This ensures the absence of minima in \textit{star worlds}.}
\label{fig:vectorfield_starshaped}
\end{figure}

The diagonal eigenvalue matrix is given as:
\begin{equation}
  \matr D(\xi) =
  \textbf{diag}
  \left(
    \lambda^r(\xi) ,
    \lambda^e(\xi) , \,
     \hdots ,
     \lambda^e( \xi)
     \right)
  \label{eq:eigVecMatr}
\end{equation}
with eigenvalues in reference direction $\lambda_r(\xi) \in ]0, 1[$ and in tangent $\lambda_e(\xi) \in ]1, 2[$ are a function of the averaged reference direction $\hat{\vect r}^o(\xi)$:
\begin{equation}
    \lambda_r(\xi) = 1 - \| \hat{\vect r}^o(\xi) \|^{\rho}
    \qquad
    \lambda_e(\xi) = 1 + \| \hat{\vect r}^o(\xi) \|^\rho
\end{equation}
where $\rho \in \mathbb{R}_{>0}$ is the reactivity, see \cite{khansari2012dynamical} for more information.\\
\\
\noindent\textbf{Theorem 1}
\textit{Consider a star-world as defined in Eq.~(\ref{eq:star_shape}) with $N^{\mathrm{obs}}$ obstacles. Any trajectory $\{\xi\}_t$, that starts within the free space $\mathcal{X}^e$, and evolves according to Eq.~(\ref{eq:modulated_ds}) and (\ref{eq:normal_vector}), will stay in free space, i.e., $\{\xi\}_t \in \mathcal{X}^e \; \forall \, t $. Furthermore, if the motion results from linear nominal dynamics $v^N$ as given in Eq.~(\ref{eq:linear_dynamics}), the final dynamics are stable and have a unique minimum at the attractor $\xi^a \in \mathcal{X}^f$, i.e., $\{ \xi : \| \dot{\xi} \| = 0, \nabla \dot{\xi} \succ 0 \} = \{ \xi^a\} $
} \hfill \textbf{Proof:}~see~Appendix~\ref{sec:proof_theorem1}.\\
\\
The proposed approach allows collision avoidance in a star-shaped world similar to \cite{huber2021avoiding}, but with only the need for a single modulation resulting in lower computational cost (see Fig.~\ref{fig:vectorfield_starshaped}).

\subsection{Tail Negligence}
The modulation-based obstacle avoidance is active even when the nominal dynamics $v^N$ are already moving away from an obstacle. This can be an undesired effect since, in this case, the unconstrained velocity already ensures collision avoidance. This \textit{tail} effect can be reduced by modifying the wake eigenvalues while still ensuring a smooth vector field by adapting the eigenvalues as follows:
\begin{align*}
  & \lambda^{e, t}(\xi) = w^r w^v + ( 1 - w^r w^v) \lambda^e(\xi) \\
  & \lambda^{r, t}(\xi) = \lambda^{e, t}(\xi) w^r \text{sgn}(w^v) + ( 1 - w^r \text{sgn}(w^v)) \lambda^r(\xi)
\end{align*}
with the weights given as
\begin{equation}
  w^r =  \min \left(1, \frac{1}{\| \hat{\vect r}^o (\xi) \|}  \right)
  \quad
  w^v =  \max \left(0, \frac{\dotprod{\vect{r}^o(\xi)}{\vect v^N}}{\|\vect v^N\|} \right)^{c^w}
\end{equation}
where $c^w \in \mathbb{R}_{>0}$ is the power weight, we choose $c^w=0.2$. 

\subsubsection{Decreasing Tail Weight}
The importance of obstacles in the wake of the nominal velocity $v^N$ can be further decreased by modifying the weight based on the nominal velocity $\vect v$. For this we restate the weight from Eq.~(\ref{eq:weight_sum_obstacle}) as:
\begin{equation*}
  \hat{w}_i^o = \hat{w}_i^o \left(\frac{\hat{w}_i^o}{\sum_i \hat{w}^o_i } \right)^{1/c_i}
  , \;\;
  c_i = 1 -  \normdotprod{\vect v^N}{\vect r_i^o(\xi)}
\end{equation*}
for all obstacles $i = 1 .. N^{\mathrm{obs}}$.

\section{Sample Based Obstacle Avoidance} \label{sec:single_modulation}
\begin{figure}[t]
\centering
\includegraphics[width=0.7\columnwidth]{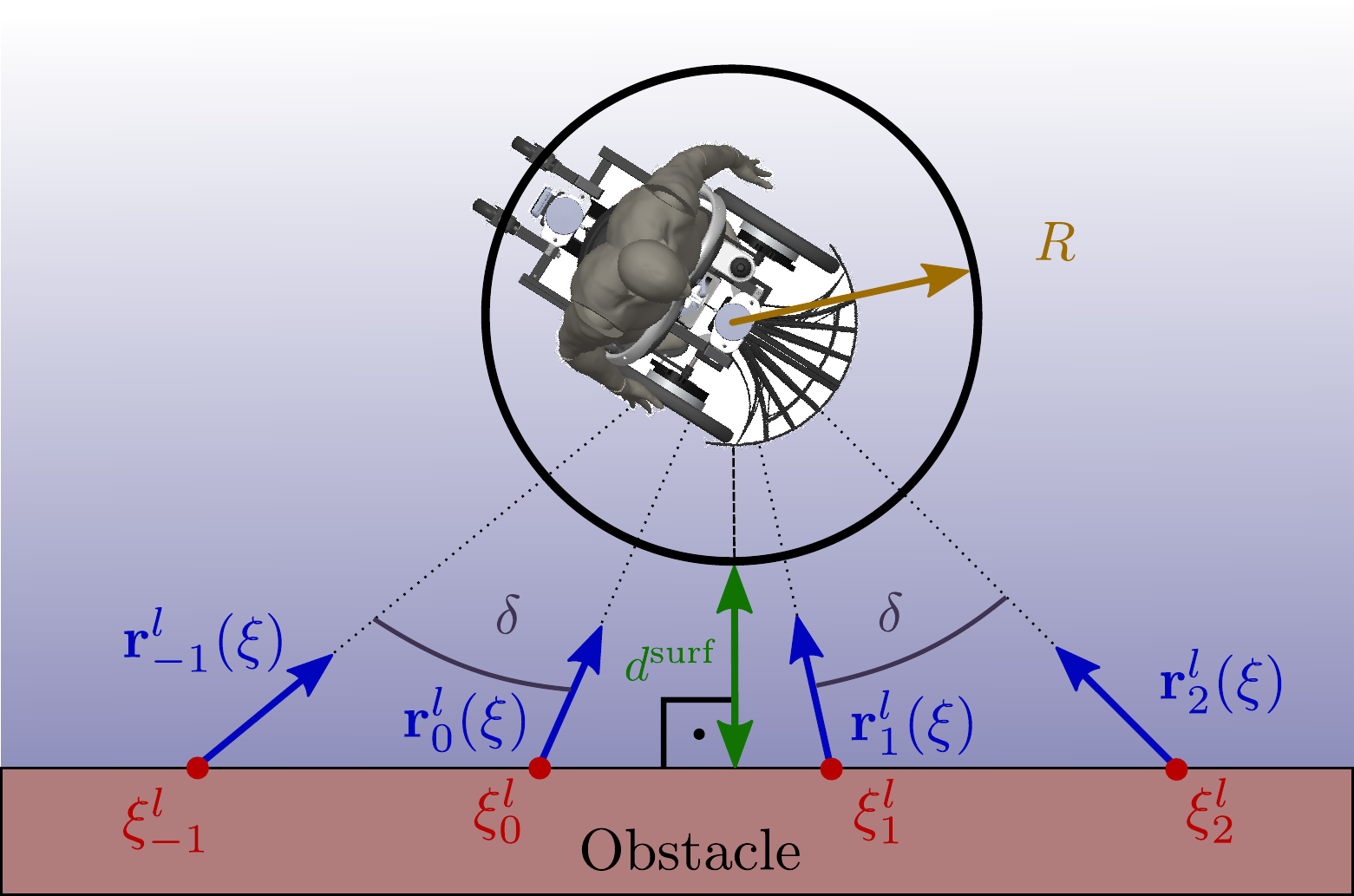}
\caption{Mobile robot in front of a flat obstacle.}
\label{fig:sensor_data_avoidance_flat}
\end{figure}

In many scenarios, the analytical obstacle description is not known at runtime, but the obstacles are perceived as a large number of data points, $\xi^p_i \in \mathbb{R}^d \; ,  i = 1, \, ..., \, N^p$, where $N^p \in \mathbb{N}_{>0}$ is the total number of points.\footnote{In this paper we focus on Lidar data, but this could be a point-cloud model of objects.}  We extend the notion of \text{virtual} obstacle to be applied to a large number of zero-dimensional data points. Furthermore, throughout this section, we assume a circular d-dimensional robot with radius $R \in \mathbb{R}_{>0}$ and a sampling angle of $\delta \ll 1$, see Fig.~\ref{fig:sensor_data_avoidance_flat}.

\subsection{Reference Summing for Sampled Data}
First, we evaluate the reference direction to each point $i$
\begin{gather}
    \hat{\vect r}^p_i (\xi) = \frac{\xi^p_i - \xi}{\| \xi^p_i - \xi \|}  
    \qquad i = 1 .. N^p
\end{gather}
Additionally, the distance is evaluated as
\begin{equation}
  D_i (\xi) = \| \xi^p_i - \xi \| - R \qquad i = 1 .. N^p \label{eq:distance_function}
\end{equation}  
This provides the reference direction similar to (\ref{eq:reference_summing}). Since each data point is approximated with a small circle, the decomposition matrix $\matr E(\xi)$ is constructed to be orthogonal. 
Furthermore, we introduce the weight sum for discrete data points:
\begin{equation}
  w^{\mathrm{sum}} = \min \left( \sum_i \hat{w}_i^p (\xi), w^{\mathrm{norm}} \right)
  \label{eq:weight_sum}
\end{equation}
where $w^{\mathrm{norm}} > 0$ is the maximum distance weight, for more information see Appendix~\ref{sec:proof_theorem1}.

\subsection{Eigenvalues Based on Reference Direction} \label{sec:eigenvalues}
The modulated velocity $\dot \xi$ needs to be tangent to the surface or point away from it when approaching the obstacle to ensure collision avoidance. Conversely, the obstacle avoidance controller should have no effect far away from any obstacle.

Correspondingly, the magnitude of the reference direction $\| \hat{\vect{r}}^p(\xi) \|$ reaches zeros far away from an obstacle, and approaches infinity when on the surface. Hence, we can use the magnitude to define the eigenvalues as follows:
\begin{align*}
  &\| \hat{\vect r}^p(\xi) \| = 0 \quad \Rightarrow \quad \lambda^r(\xi) = \lambda^e(\xi) = 1 \\
  &\| \hat{\vect r}^p(\xi) \| = 1 \quad \Rightarrow \quad  \lambda^r(\xi) / \lambda^e(\xi) = 0 \\
  &\| \hat{\vect r}^p(\xi) \| \rightarrow \infty  \quad \Rightarrow \quad | \lambda^e(\xi) |/ \lambda^r(\xi) = 0
\end{align*}
Additionally, we set the constraint that $\lambda^e(\xi) \in \mathbb{R}_+$, since the velocity in the tangent direction should never be reflected.
We choose the eigenvalue in reference direction $\lambda^r(\xi, \vect v) \in [1, \, -1[$ as:
\begin{align}
  & \lambda^r(\xi, \vect v) =
  \begin{cases}
    (-1) \lambda^{r, 0}(\xi) \quad & \text{if} \;\;  \dotprod{\vect r^p(\xi)}{\vect{v}}  < 0 , \; \| \hat{\vect r}^p(\xi) \| > 1 \;\; \\
    \lambda^{r, 0}(\xi) & \text{otherwise}
  \end{cases} \nonumber \\
  & \quad \text{with} \;\;
  \lambda^{r, 0}(\xi) =
  \begin{cases}
    \cos( \frac{\pi}{2} \| \hat{\vect r}^p(\xi) \|) & \text{if} \;\; \| \hat{\vect r}^p(\xi) \| < 2 \\
    -1 & \text{otherwise}
    \label{eq:eigenvalues_simples}
  \end{cases}
\end{align}
Additionally the eigenvalues in tangent direction $\lambda^r(\xi) \in  [1, \, 2]$ are given by
\begin{equation}
  \lambda^e(\xi) =
  \begin{cases}
    1 + \sin( \frac{\pi}{2} \| \hat{\vect r}^p(\xi) \|) & \text{if} \;\; \| \hat{\vect r}^p(\xi) \| < 1 \\
    2 \sin \left( \frac{\pi}{2 \| \hat{\vect r}^p(\xi) \|} \right) & \text{otherwise} \label{eq:eigenvalues_tangent}
  \end{cases}
\end{equation}
where $\pi$ is the circle constant. 
Note that the eigenvalues are $C^1$-smooth, because of $\frac{d}{d x} \sin(x) |_{x = \pi/2} = 0$  and $\frac{d}{d x} \cos(x) |_{x = \pi} = 0$.  The resulting vector field can be seen in Fig.~\ref{fig:motion_laserscan}.\\
\\
\noindent\textbf{Theorem 2}
\textit{Consider an agent with radius $R$ and a nominal velocity of $\vect{v}^N \in \mathbb{R}^d \setminus \vect 0$. Let us assume the existence of $N^{\mathrm{obs}}$ obstacles with corresponding boundary and free space as given in Eq.~(\ref{eq:gamma_function}), from which the sampled surface points $\xi_i^p \in \mathbb{R}^d, \, i \in 1, \, ... \, N^{p}$ are known. \\
Any trajectory $\{\xi\}_t$ that starts in free space $\mathcal{X}$, and evolves according to  Eq.~(\ref{eq:modulated_ds}), (\ref{eq:eigenvalues_simples}) and (\ref{eq:eigenvalues_tangent}) will stay in free space, i.e., $\{\xi\}_t \in \mathcal{X}^e$ 
and will not stop outside of the gap distance, i.e., $\| \dot{\xi} \| > 0 \;\; \forall \xi \in \mathcal{X}^g$.
}
${ \, }$ \hfill \textbf{Proof:}~see~Appendix~\ref{sec:proof_theorem2}. 


\begin{figure}[t]\centering
\begin{subfigure}{0.49\columnwidth} %
\centering
\includegraphics[width=1.0\columnwidth]{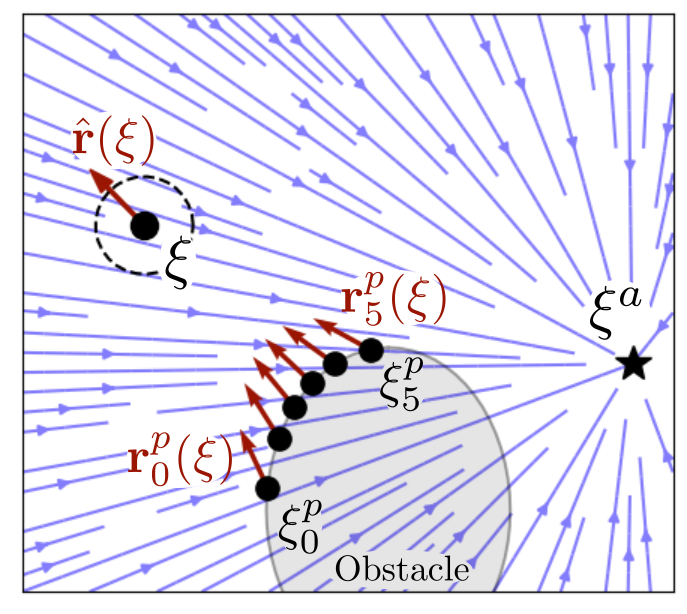}
\caption{Initial vector field}
\label{fig:summation_of_reference}
\end{subfigure}  
\begin{subfigure}{0.49\columnwidth}
\centering
\includegraphics[width=1.0\columnwidth]{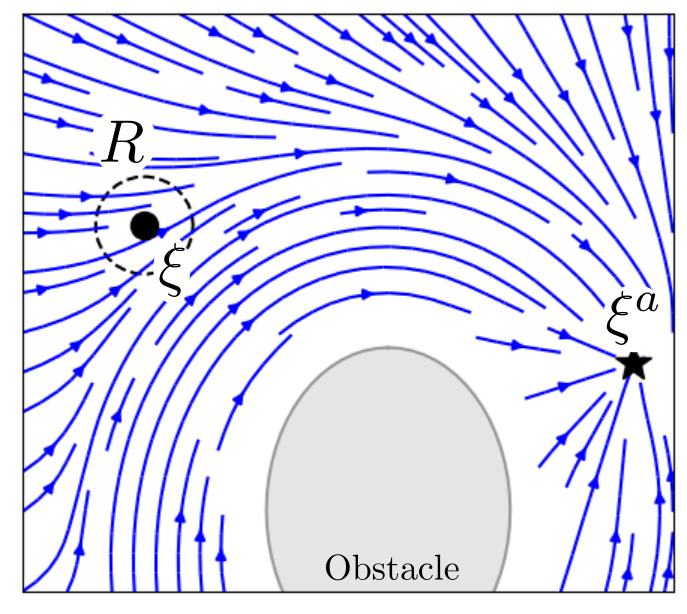}
\caption{Modulated vector field} 
\label{fig:avoided_vectorfield}
\end{subfigure}%
\caption{At each position, the algorithm is only aware of the sample points $\xi^p_i, \; i  = 1 .. N^p$ and not the actual shape of the obstacle in grey (a). The the summed reference direction $\hat{\vect r}^p(\xi)$ is used to guide the nominal velocity around the obstacle in grey (b).}
\label{fig:motion_laserscan}
\end{figure}

\begin{figure}[t]
\centering
\includegraphics[width=0.9\columnwidth]{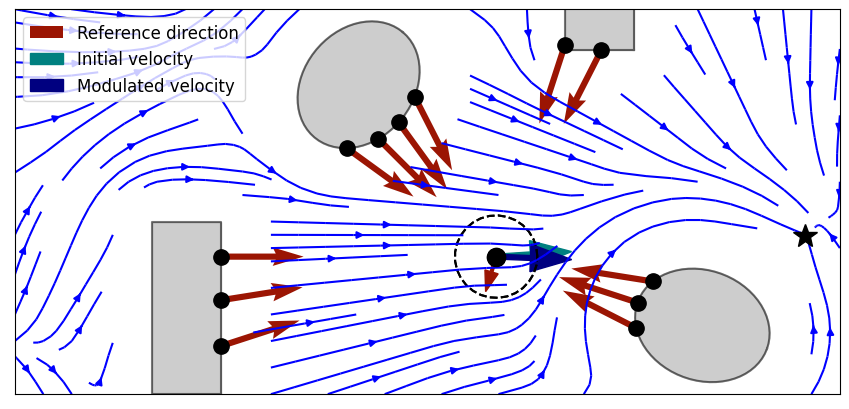}
\caption{The algorithm does not need to estimate the number or shape of each individual obstacle. Calculating the weighed reference direction is sufficient to navigate safely in cluttered environments.}
\label{fig:motion_laserscan_multiple}
\end{figure}

\section{Disparate Obstacle Descriptions} \label{sec:mixed_environments}
Detection algorithms can fail to identify certain obstacles. Hence, we are left with a mix of analytic obstacles and sampled sensor data. Furthermore, the analytic reconstruction of the obstacles often comes with a significant delay in time.
This results in asynchronous reception of information on obstacles, with fast sampled-based information and slow analytical model acquisition.
We address this by proposing an approach to fuse analytic obstacle descriptions (see Sec.~\ref{sec:non_circular}) with sample-based sensor  data (see Sec.~\ref{sec:single_modulation}).

The first step is to remove sampled obstacle data which has been identified as an analytic obstacle. Hence, the laserscan weights in (\ref{eq:weight_sum}) are set to zero if they lie withing the obstacle $o = 1 .. N^{\mathrm{obs}}$:
\begin{equation}
  \hat{w}^o_i(\xi)
    \gets 0 \qquad  \text{if} \;\; \exists \, o : \, \Gamma_o(\xi_i)  \leq 1
\end{equation}

\subsection{Fusion Weights of Sampled and Analytic Obstacles}
The magnitude of the reference direction of the sampled and analytic obstacles gives us information about the proximity to obstacles. 
Hence, the reference magnitude can be used to calculate the importance weight of the sampled data $w^p \in [0, 1]$ and analytic obstacles $w^o \in [0, 1]$ as:
\begin{equation*}
  \begin{bmatrix} w^p\\ w^o \end{bmatrix}
  = \frac{1}{\hat w^p  + \hat w^o} \begin{bmatrix} \hat w^p \\ \hat w^o \end{bmatrix}
  \quad \text{with}  \;\; 
  \begin{cases} 
  \hat w^p = 1 / \left( 1 - \left\| \bar{\vect r}^p(\xi) \right\|  \right) - 1 \\
  \hat w^o = 1 / \left( 1 - \left\| \bar{\vect r}^o(\xi) \right\|  \right) - 1 
  \end{cases}
\end{equation*}
where $i = 1 .. N^{\mathrm{obs}}$. The weight is used to evaluate the reference direction of the mixed environment:
\begin{equation}
  \vect{r}^m(\xi) = w^l(\xi)  \hat{\vect{r}}^p(\xi) + w^o(\xi) \hat{\vect{r}}^o(\xi)
\end{equation}


The analytical description has the additional advantage that it often provides information about the velocity and shape of the obstacles. This can be used to ensure collision avoidance in dynamic environments. We propose to include the velocity similar to \cite{huber2021avoiding} for the fused scenario as follows:
\begin{equation}
  \dot{\vect \xi} = { \matr{M}(\xi, \vect v^N)} \left( \vect v^N - \dot{\tilde \xi}^{\mathrm{d, \mathrm{tot}}} \right) + \dot{\tilde \xi}^{d, \mathrm{tot}}
  \; \, \quad
  \dot{\tilde \xi}^{d, \mathrm{tot}} = w^o \dot{\tilde \xi}^{\mathrm{tot}}
\end{equation}
where $\dot{\tilde \xi}^{\mathrm{tot}} \in \mathbb{R}^d$ is the averaged velocity of all analytical obstacles.

\begin{figure}[t]
\centering
\includegraphics[width=1.0\columnwidth]{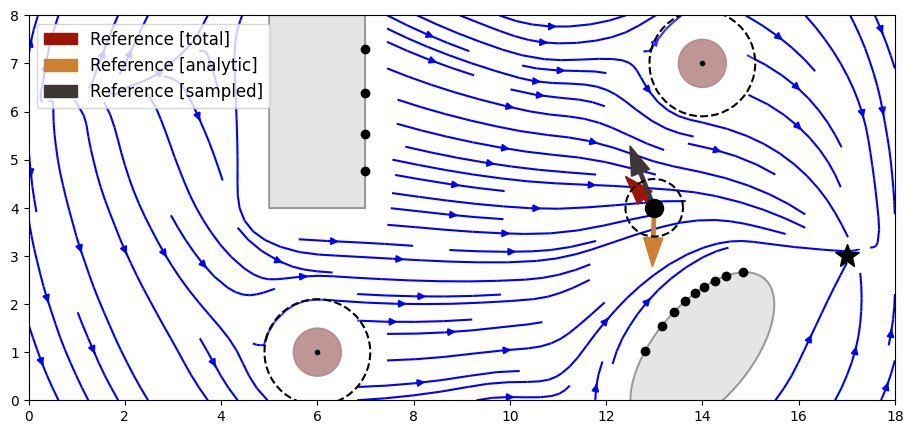}
\caption{Analytical obstacle descriptions (brown) and sampled obstacle data (black dots on the gray obstacles) are fused by the algorithm. The weighting uses the corresponding reference directions.}
\label{fig:mixed_streamplot}
\end{figure}

\subsection{Asynchronous Evaluation}
\begin{figure}[tb]
\centering
\includegraphics[width=0.91\columnwidth]{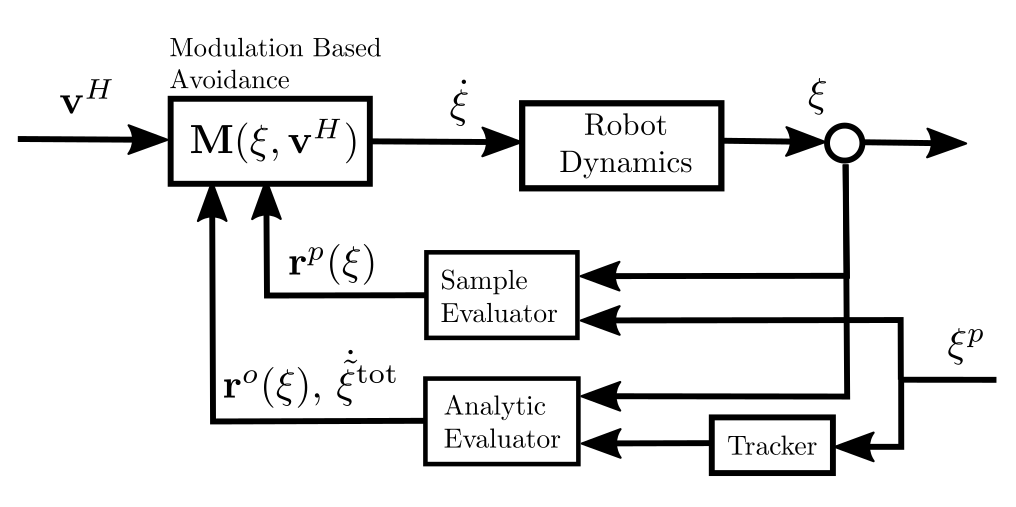}
\caption{The modulation-based avoidance takes the most recent update from the sampled and analytic obstacle description. The reference directions are obtained at different frequencies and are combined asynchronously.}
\label{fig:control_scheme_mixed}
\end{figure}

The disparate data is combined by using a controller with three inputs: the velocity command from the human operator $\vect{v}^N$, the laser scan reading  $\xi^p$, and the estimated position of the robot $\xi$  (see Fig.~\ref{fig:control_scheme_mixed}). 

The two sensor loops run at different frequencies, with the sample-based evaluator running at a higher update rate than the analytic evaluator. The latter is delayed due to the processing time to generate analytic obstacle descriptions. However, the analytic descriptions enable improved behavior due to the information about the obstacle's shape and velocity.

\subsection{Uniform Importance Scaling} \label{sec:importance_scaling}
An obstacle described by sampled data points or an analytic function should have the same effect on the modulation. An adequate scaling is necessary since a single obstacle often corresponds to multiple data points.

Let us consider an obstacle that can be approximated by radius $R^{\mathrm{obs}}$ at a distance $D$. The number of sampling data points $N^p$  can be approximated as:\footnote{If the obstacle sampling is volumetric, e.g., by obtaining a collision model of the robot in higher dimensions, the power value of the right side is $d$, i.e., $N^p \sim (\cdot)^d$.}
\begin{equation}
N^p 
\sim \left(\frac{R^{\mathrm{obs}}}{\tan(\delta) D}\right)^{d-1}
\sim \left(\frac{R^{\mathrm{obs}}}{\delta D}\right)^{d-1}
\end{equation}

It follows that the scaling weight of the sampled data is set proportional to the sampling angle $\delta$:
\begin{equation}
    w^{\mathrm{scal}, p} \approx  \delta ^{d-1}
\end{equation}

Moreover, the scaling weight of the analytic obstacle description is dependent on the size and distance.
\begin{equation}
    w^{\mathrm{scal}, o} \approx  \left( R^{\mathrm{obs}} / D\right)^{d-1}
\end{equation}
As a result, we set the distance scaling
$D^{\mathrm{scal}}= 2 \pi / \delta^{(d-1)}$ in mixed environments.

\section{Experimental Validation}
\subsection{Computational Speed}
\begin{figure}[t]\centering
\begin{subfigure}{0.49\columnwidth}
\centering
\includegraphics[width=1.0\columnwidth]{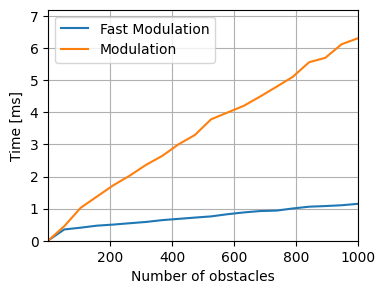}
\caption{Analytic data} 
\label{fig:comparison_algorithms}
\end{subfigure}%
\begin{subfigure}{0.5\columnwidth} %
\centering
\includegraphics[width=1.0\columnwidth]{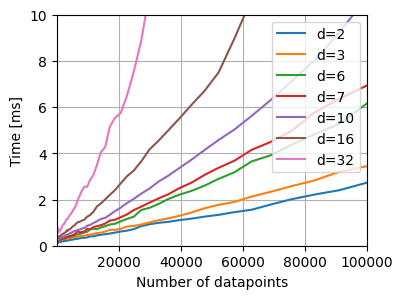}
\caption{Sample data}
\label{fig:comparison_sampling_dimensions}
\end{subfigure}  
\caption{The proposed approach (\textit{Fast Modulation}) and the baseline \cite{huber2019avoidance} denoted as \textit{Modulation} have a computation time that grow linearly with the number of obstacles. However, the proposed approach has a flatter rate of growth (a). Moreover, the complexity grows linearly with respect to the number of data points (b). The evaluation was performed on a desktop computer with \textit{an Intel Core i7-6700 CPU @ 3.40GHz}.}
\label{fig:computational_speed_comparsion}
\end{figure}

The proposed, fast obstacle avoidance algorithm has a computational complexity of $\mathcal{O} (d N^{\mathrm{obs}})$, is lower than when compared to similar approaches that have a complexity of $\mathcal{O} (d^{2.4} N^{\mathrm{obs}})$, see \cite{huber2021avoiding}. This was confirmed by the low computational time observed during the experiments for sampled data and analytic obstacle descriptions (see Fig.~\ref{fig:computational_speed_comparsion}).  \\
Note that an important speed-up of the presented algorithm compared to the baseline method is due to the application using sampled sensor data directly without using an intermediate tracker.
 
\subsection{Convergence Analysis Using Disparate Sensor Data}
\begin{figure}[t]\centering
\begin{subfigure}{0.49\columnwidth}
\centering
\includegraphics[width=1.0\columnwidth]{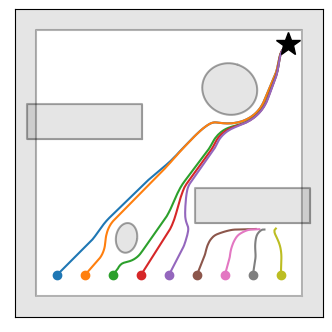}
\caption{Sampled data}
\label{fig:sampled_data_comparison}
\end{subfigure}%
\begin{subfigure}{0.5\columnwidth} %
\centering
\includegraphics[width=1.0\columnwidth]{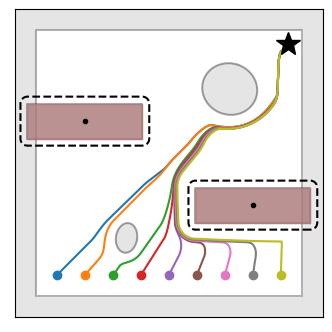}
\caption{Disparate data}
\label{fig:disparate_data_comparison}
\end{subfigure}  
\caption{Trajectories often end in a local minimum when using sampled data of obstacles only, see (a). The avoidance can be improved using analytical obstacle description (brown), where an increased number of trajectories converge to the attractor (black star) as in (b).}
\label{fig:comparsion_two_partial_algorithms}
\end{figure}

The algorithm was applied to simulated, two-dimensional environments with four obstacles and a surrounding wall (Fig.~\ref{fig:comparsion_two_partial_algorithms}). The shape and pose of the ellipse obstacles were randomly obtained in the top right and bottom left corner, respectively. The position of the two square obstacles is fixed, but the obstacle is either discovered through surface sampling or using the analytical obstacle description. The surrounding wall is fixed. 

From the comparison between the two scenarios, it can be observed that the analytical knowledge of the obstacles doubles the number of converging trajectories, see Tab.~\ref{tab:comparison_convergence}. 
However, this increases the time to convergence by around 20 \%, resulting from a more conservative behavior around analytic obstacles. 

\begin{table}[t]
\caption{The ratio of trajectories (100 runs) that reached the attractor for implementation with only sampled data versus the use of disparate environment description. The average time is taken over the runs which had both algorithms converging.}
    \centering
    \begin{tabular}{|l|c|c|} \hline
     & Sampled & Disparate \\ \hline
    Convergence Ratio & 45\% & 83\% \\
    Average Time & \SI{ 24.6 }{s} & \SI{29.7}{s}  \\  
    \hline
    \end{tabular}
    \label{tab:comparison_convergence}
\end{table}

\subsection{Evaluation using Static Sensor Data}
An additional evaluation was performed with two static recordings of real sensor data, which was obtained by two 3D \textit{Velodyne} Lidars. Each Lidar is interpreted as a laser scan by extracting the horizontal 2D row of the Lidar. During the recording, the sensor was placed approximately \SI{30}{cm} above ground. The resulting laser scans have an angle of view of $\pm 0.75 \, \pi \, \mathrm{rad}$ and an angle increment of $\delta=$\SI{7e-3}{rad}, resulting in approximately 650 data points each.

\begin{figure}[t]\centering
\begin{subfigure}{0.8\columnwidth}
\centering
\includegraphics[width=1.0\columnwidth]{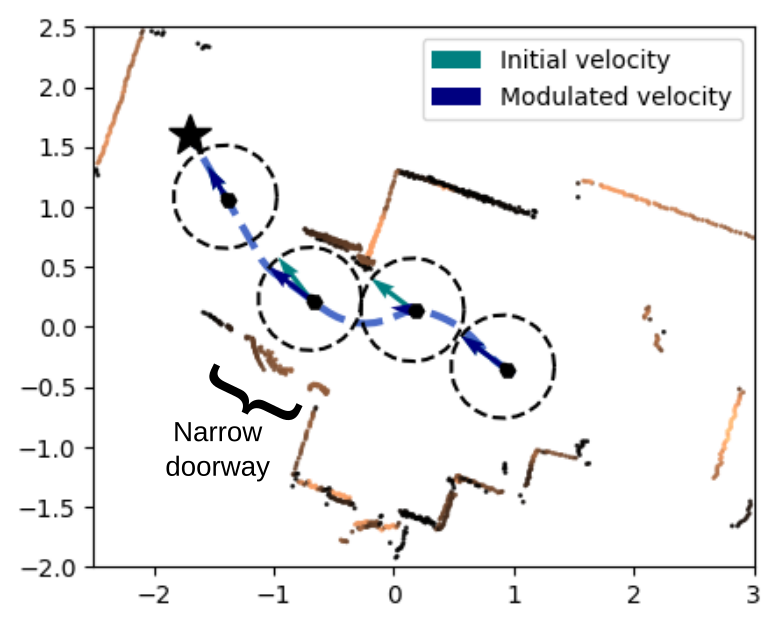}
\caption{Circular robot passing a narrow doorway}
\label{fig:passing_doorway}
\end{subfigure}
\begin{subfigure}{0.8\columnwidth} %
\centering
\includegraphics[width=1.0\columnwidth]{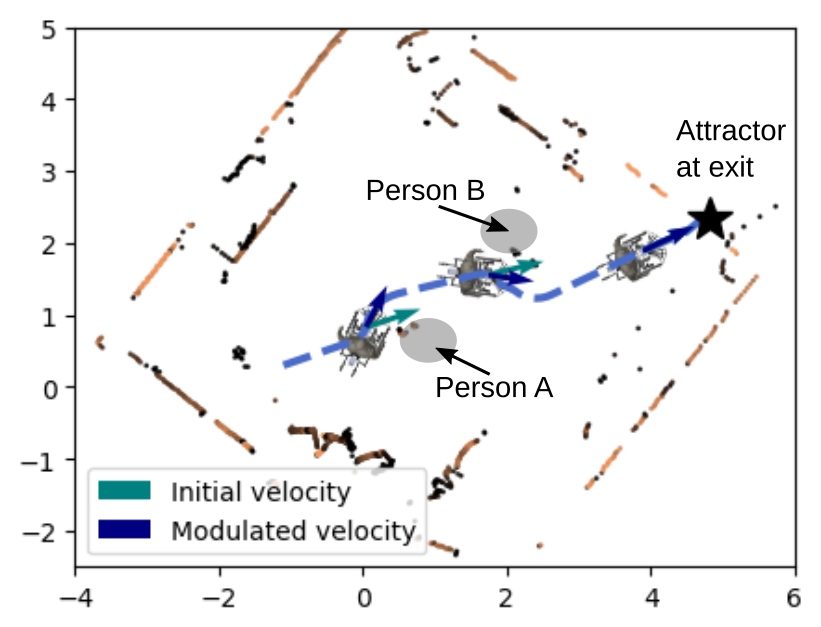}
\caption{QOLO passes a room with two people}
\label{fig:comparison_sampling_dimensions}
\end{subfigure}  
\caption{In experiments using instances of recorded sensor data of approximately 1000 data points, the robot is able to navigate between the narrow doorways (a) and cluttered environments (b).}
\label{fig:motion_laserscan}
\end{figure}

Based on the static laser scan, the robot passes narrow doorways successfully (Fig.~\ref{fig:passing_doorway}). Even though the gap is marginally wider than the (simulated) robot's diameter, no local minimum is observed in front of the doorway. 
When passing a room, the simulated robot avoids obstacles of various shapes in real-time. The two people standing in the room were avoided based on sampled data only (Fig.~\ref{fig:comparison_sampling_dimensions}).

\subsection{Outdoor Crowd Evaluation with QOLO} \label{eq:evaluation_qolo}
The experimental validation is performed with the semi-autonomous, standing wheelchair QOLO \cite{granados2018unpowered}. The nominal velocity $\vect v^N$ was obtained from the onboard operator. 
The algorithm was implemented in \textit{Python} and evaluated on the onboard computer of QOLO (\textit{Intel(R) Atom(TM) Processor E3950 @ 1.60GHz}), taking \SI{0.4}{ms} to evaluate a scenario with over 1000 data points. Furthermore, the main control loop runs at \SI{100}{Hz}, with Lidar information being obtained at \SI{20}{Hz} and the pedestrian-tracker providing an update at a frequency of \SIrange{1}{5}{Hz} (depending on the number of tracked persons in close proximity). The operator control input was obtained at a frequency of \SI{20}{Hz}.

The wheelchair is non-holonomic with two rigid main wheels and two passive back wheels; its geometry is approximated with a circle of radius \SI{0.45}{m}. The circle's center $\xi$ is placed in front of the middle of the main wheels at a distance of $d^c=$~\SI{6.25e-2}{m}. This is the position where the dynamical system is also evaluated. 
The modulated velocity $\dot \xi$ is split into a linear command $\dot{\xi}_l$ and angular command $\dot{\xi}_a$ by using the Jacobian $J^Q$:
\begin{equation}
    \begin{bmatrix} \dot{\xi}_l \\ \dot{\xi}_a \end{bmatrix} 
    = 
    \left( J^Q \right)^{-1} \dot{\xi}
    \qquad \text{with} \quad
    J^Q = \begin{bmatrix} 1 & 0 \\0 & d^c  \end{bmatrix} 
\end{equation}



\begin{figure}[tb]
\begin{subfigure}{0.495\columnwidth} %
\centering
\includegraphics[trim={12cm 2cm 15cm 0}, clip, width=1.0\columnwidth]{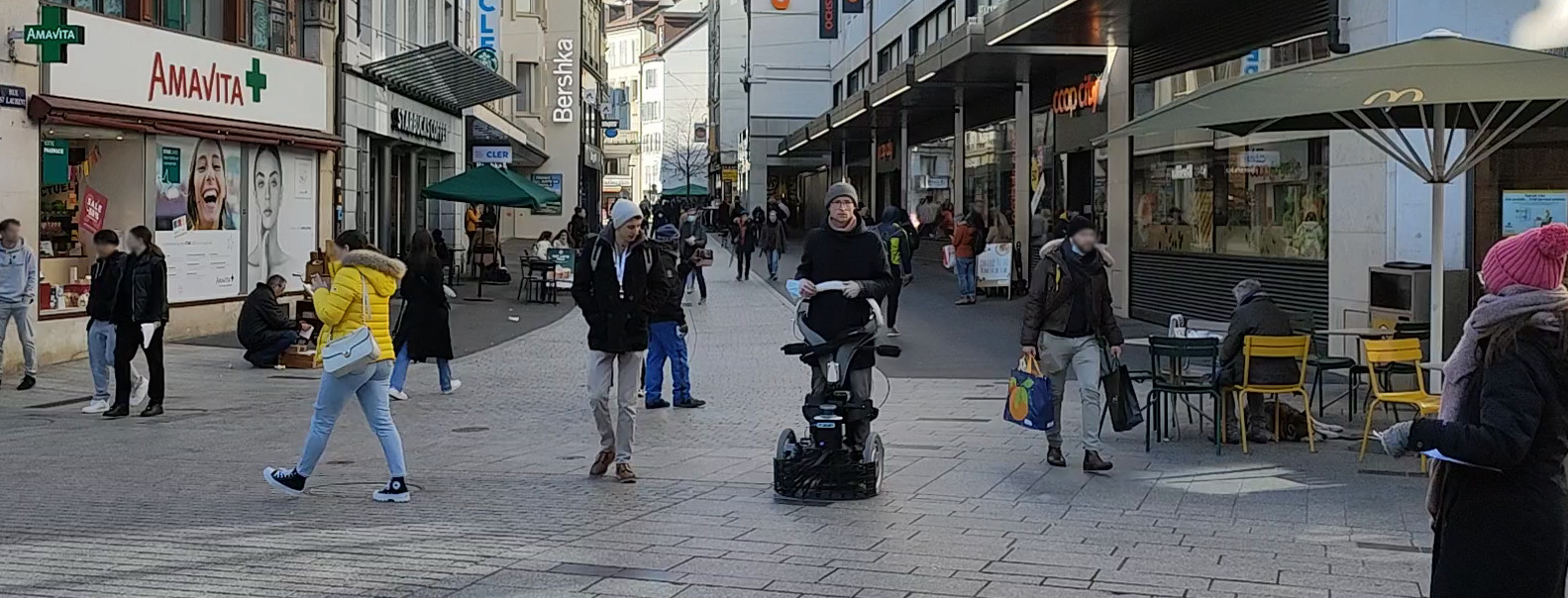}
\caption{QOLO in the city center}
\label{fig:qolo_with_david}
\end{subfigure} %
\begin{subfigure}{0.495\columnwidth} %
\centering
\includegraphics[trim={0 0.45cm 0cm 0}, clip, width=1.0\columnwidth]{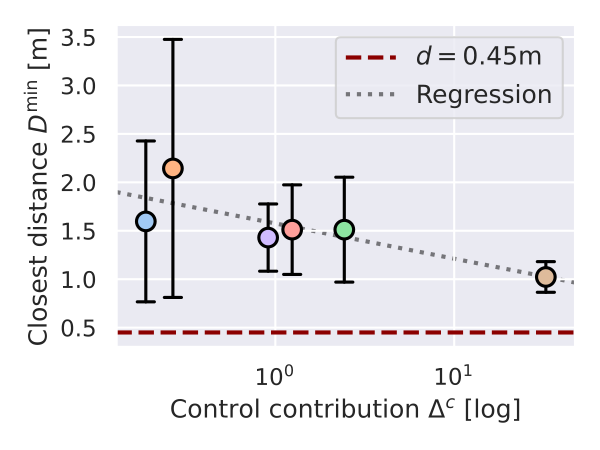}
\caption{Shared control analysis}
\label{fig:comparison_distance_control}
\end{subfigure}
\caption{During the evaluation with the QOLO robot in Lausanne (a) the control contribution ensured that the closest distance to any pedestrian stayed above the save margin of \SI{0.45}{m} during six runs (b). The control contribution is the relative input of the controller, evaluated as $\Delta ^c = \| \dot{\xi} - \vect v^N \| / \| \vect v^N \|$. The closest distance $D^{\mathrm{min}}$ is the average distance of the ten closest Lidar measurements. The relation between the mean of the two measurements was observed to be $\Delta ^c = \exp{(10.0 - 6.3 D^{\mathrm{min}})}$.}
\label{fig:qolo_in_lausanne}
\end{figure}

The qualitative evaluation with the robot was conducted in the city center of Lausanne, Switzerland.\footnote{Approval has been obtained from the EPFL ethics board and the city of Lausanne. The driver was on board the robot during the experiment and could interfere. Two experimenters watched the scene and verified the controller output.}
The chosen location is an intersection of six streets and is restricted to pedestrians. This results in a large diversity in pedestrian speeds and directions of movement. The operator was navigating with QOLO up and down the street along a transect with a distance of \SI{15}{meters} (Fig.~\ref{fig:qolo_in_lausanne}). During the experiments, the controller supported the operator while ensuring a collision-free motion. 



\section{Discussion}
This work presents a modulation-based approach for real-time obstacle avoidance based on dynamical systems. The proposed method excels in its speed and can be applied to sampled data, such as Lidar or laser scans combined with analytic obstacle descriptions. 
We proved the absence of local minima in free space, and an agent controlled with our algorithm only stopped when directly driving towards obstacles. Furthermore, the method is scalable to higher dimensions. 
Experimental validation was performed in simulation, static sensor data, and the standing wheelchair QOLO. It was shown that the algorithm avoids collisions while navigating in static and dynamic environments. 
The evaluation with sampled sensor data showed that the algorithm could pass narrow doorways. %
Moreover, the algorithm had a short evaluation time onboard the robot.

\subsection{Future Work}
Most theoretical development in this work is for multiple-dimensional collision avoidance. However, the implementation is on a mobile agent in crowds for navigation in two-dimensions. Future implementations will focus on exploiting the scalability to higher dimensions. Sampled data is obtained through distance sensors for two and three-dimensional cases. Conversely, in higher dimensional space, e.g., joint space, sampled data about the environment can be obtained through sampling in simulations \cite{mirrazavi2018unified, koptev2021real}. 

The algorithm has been tested in a shared-controller setup. Future work will include a quantitative controller evaluation compared to existing algorithms in various crowd scenarios. Additionally, this will include several operators and see how different driving styles influence the overall behavior. Finally, we believe that quantitative data of the algorithm will enable the design of an intelligent shared-controller, which learns and adapts its behavior based on the operator and the environment, similar to \cite{li2011dynamic}.

\section{Acknowledgement}
We want to thank David Gonon and Albéric de Lajarte for their support in running the experiments on QOLO.

\appendix

\subsection{Proof of Theorem 1} \label{sec:proof_theorem1}
As the eigenvalues $\lambda^e(\xi)$ are equal in all tangent directions $\vect e_i$, we only analyze the two-dimensional case. Generalization is the immediate result; see \cite{huber2019avoidance} for further elaboration.

\subsubsection{Absence of Local Minima in Free Space}
Les us first show that the dot-product is uniquely positive, and hence describes a (virtual) star shaped obstacle as defined in (\ref{eq:star_shape}), 
\begin{align*}
  \vect{r}(\xi)^T \hat{\vect{n}}(\xi)
  = \vect{r}(\xi)^T (c^n \vect{r}(\xi) +  {\vect n}^{\Delta}(\xi))
  = c^n + \vect{r}(\xi)^T {\vect n}^{\Delta}(\xi) 
\end{align*}
Further, from the star-shape constraint in (\ref{eq:star_shape}), we have $\|\vect n^{\Delta}(\xi)\| < \sqrt{2}$.
Consider the case when $\dotprod{\vect{r}(\xi)}{\vect n^{\Delta}(\xi)} / \|\vect n^{\Delta}(\xi)\| > -\sqrt{2}/2$:
\begin{equation*}
 \vect{r}(\xi)^T \hat{\vect{n}}(\xi) = 1 + \vect{r}(\xi)^T {\vect n}^{\Delta}(\xi)  \geq 1 - \|\vect n^{\Delta}(\xi)\| \sqrt{2}/2 > 0
\end{equation*}
For the case of $\dotprod{\vect{r}(\xi)}{\vect n^{\Delta}(\xi)} / \|\vect n^{\Delta}(\xi)\| \leq -\sqrt{2}/2$ we have:
\begin{align*}
\vect{r}(\xi)^T \hat{\vect{n}}(\xi)
& = - \sqrt{2} \vect{r}(\xi)^T  {\vect n}^{\Delta} + \vect{r}(\xi)^T {\vect n}^{\Delta}(\xi) \\
& = (\sqrt{2}  - 1 ) \left( - \vect{r}(\xi)^T  {\vect n}^{\Delta}(\xi) \right) \\
& \geq (\sqrt{2}  - 1) \|\vect n^{\Delta}(\xi)\| \sqrt{2}/2 > 0
\end{align*}
Hence, in both scenarios, the dot-product is positive.


\subsubsection{Saddle Point Only on the Obstacles' Surfaces}
The avoidance of multiple obstacles is defined as the modulation with respect to a single \textit{virtual} obstacle with boundary function $\hat{\Gamma}(\xi)$. As we approach an obstacle $i$, we have $\hat{\Gamma}(\xi) \rightarrow \Gamma_i(\xi)$ (see Sec.~\ref{sec:single_modulation_avoidance}). \\
From this, we can conclude that (1) since there is no minimum on the surface of the obstacle $i$, there are no minima for the virtual obstacle, and (2) as there is impenetrability of the obstacle $i$, any trajectory starting in free space $\mathcal{X}^e$ will remain in free space. For further information, see \cite{huber2019avoidance}. 

Hence, full impenetrability and a minimum-free space follow. 
${ }$ \hfill $\blacksquare$

\subsection{Proof of Theorem 2} \label{sec:proof_theorem2}
Analogously to the proof in Appendix~\ref{sec:proof_theorem1}, the two-dimensional analysis generalizes to higher dimensions. 

\subsubsection{Continuously Defined Vector field}
Special consideration has to be given to the case when $\hat{\vect r}(\xi) \rightarrow 0$, since the vector is normalized before being used in the modulation (\ref{eq:modulated_ds}).
For $\| \hat{\vect r}(\xi) \| = 0$, the final velocity is obtained as:
\begin{align*}
  \dot \xi &= \matr E(\xi) \matr D(\xi, \vect v^N) \matr E(\xi) ^T \vect v^N \\
  & = \matr E(\xi) \text{diag}\left(\lambda^r(\xi, \, \vect v^N \right) , \lambda^e(\xi)) \matr E(\xi) ^T \vect v^N \\
  & = \matr E(\xi) \text{diag}\left( (1 + \sin(0)), \, \cos(0)  \right) \matr E(\xi) ^T \vect v^N \\ 
  & = \matr E(\xi) \, \matr I \, \matr E(\xi) ^T \vect v^N = \vect v^N
\end{align*}
Additionally, all variables are continuously defined; hence the modulation effect is low in the surrounding region, and the DS is continuously defined.

\subsubsection{Absence of Local Minima in Free Space}
The modulated DS from Eq.~(\ref{eq:modulated_ds}) has the same equilibrium points as the original system, if $\det \left( \matr M(\xi) \right) \neq 0$, see \cite{kronander2015incremental} for further information. 
Since the basis matrix $\matr E(\xi)$ is orthonormal, the modulation matrix is only \textit{rank-deficient} if any of the eigenvalues are 0. From Eq.~(\ref{eq:eigenvalues_simples}), we know that the only trivial eigenvalues are at $\lambda^e(\xi) |_{\|\hat{\vect r}^p(\xi) \| \rightarrow \infty} = 0$ and $\lambda^r(\xi) |_{\|\hat{\vect r}^p(\xi)\| = 1}  = 0$. 

\begin{figure}[t]\centering
\begin{subfigure}{0.32\columnwidth} %
\centering
\includegraphics[width=1.0\columnwidth]{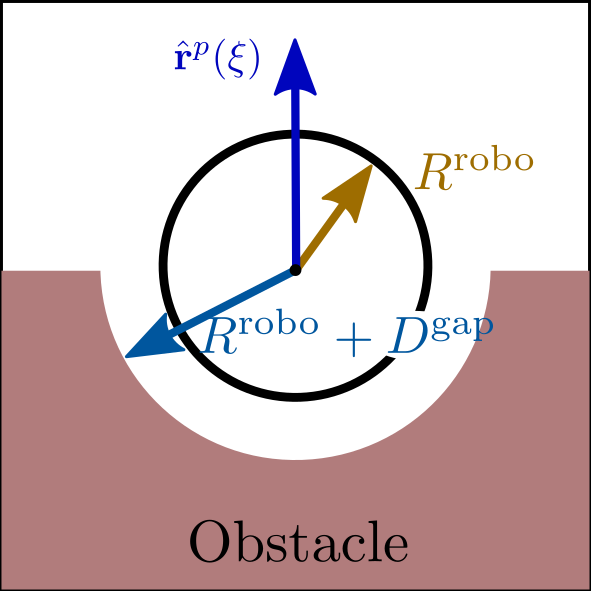}
\caption{Repulsive gap}
\label{fig:repulsive_gap}
\end{subfigure} %
\hfill
\begin{subfigure}{0.32\columnwidth}
\centering
\includegraphics[width=1.0\columnwidth]{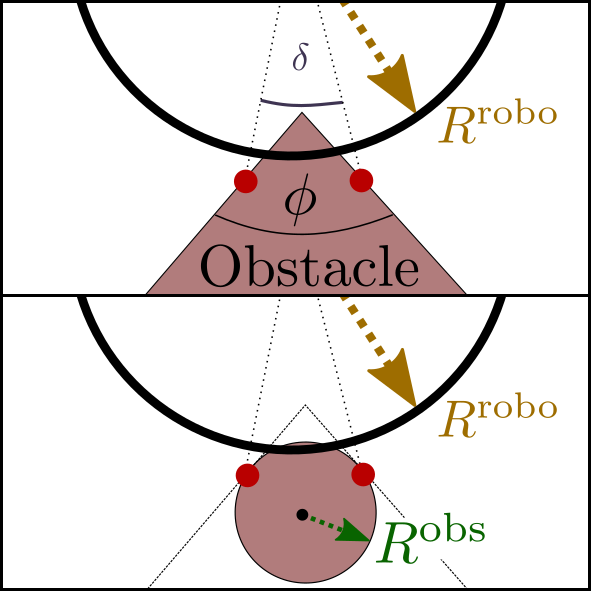}
\caption{Small obstacles} 
\label{fig:small_obstacles}
\end{subfigure}%
\hfill
\begin{subfigure}{0.32\columnwidth}
\centering
\includegraphics[width=1.0\columnwidth]{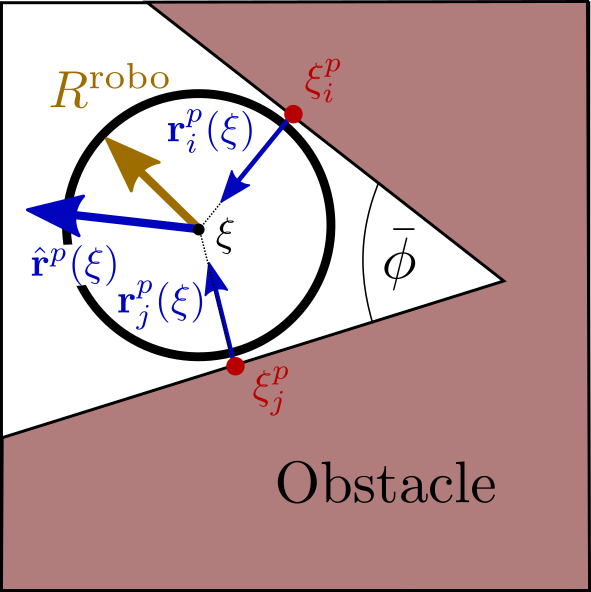}
\caption{Concave regions} 
\label{fig:concave_region}
\end{subfigure}
\caption{Multiple extreme environments for the sensor reading.}
\label{fig:extreme_sensor_situations}
\end{figure}

It is hence sufficient to limit the magnitude of the reference direction as $\| \hat{\vect r}^p(\xi) \|  < 1$. Let us analyze the \textit{maximum repulsive environment} as drawn in Fig.~\ref{fig:repulsive_gap}), i.e., where the distance to the surface $D_i^p$ in one half-plane is $D^{\mathrm{gap}}$, and infinite in the other half-plane:
\begin{equation*}
  D_i^p =
  \begin{cases}
    D^{\mathrm{gap}} & \text{if} \;\; \left(\hat{\vect r}^p\right)^T \cdot \xi_i^p < 0 \\
    \infty & \text{otherwise}
  \end{cases}
\end{equation*}
For this scene, the magnitude of the reference direction is evaluated as:
\begin{align*}
\| \hat{\vect r}^p (\xi) \|
& \leq \sum_{i = 0}^{N/2} \frac{D^{\mathrm{scal}}}{D^{\mathrm{gap}}} w^{\mathrm{sum}}\sin(i \delta)  
\approx   w^{\mathrm{sum}} \frac{D^{\mathrm{scal}}}{D^{\mathrm{gap}}} \int_0^{\pi / \delta} \sin(\phi \delta) \, d\phi \\
& =  w^{\mathrm{sum}}  \frac{D^{\mathrm{scal}}}{D^{\mathrm{gap}}} \frac{2}{\delta} \leq w^{\mathrm{norm}}  \frac{D^{\mathrm{scal}}}{D^{\mathrm{gap}}} \frac{2}{\delta}
\end{align*}
The approximation of the finite sum as an integral holds for a small sampling angle $\delta$ and many sample points, as is the case for most scanners. 

By choosing $w^{\mathrm{norm}} = (D^{\mathrm{gap}} \delta) / (2 D^{\mathrm{scal}})$, the modulated DS does not vanish further than $D^{\mathrm{gap}}$ away from any obstacle.


\subsubsection{Impenetrability}
Let us observe the modulated velocity $\dot{\xi}$ as we approach the surface, i.e., $\exists i \, : D_i^p \rightarrow 0$ (see Fig.~\ref{fig:sensor_data_avoidance_flat}), it follows that $\| \vect{\hat r}^p(\xi) \| \rightarrow \infty$ and hence $\lambda^e(\xi) = 2 \sin \left( \pi / (2 \| \hat{\vect r}^p(\xi) \|) \right) = 0$. With this, we can rewrite the modulated velocity as
\begin{align*}
  \dot{\xi} 
  & =  \matr E(\xi) \text{diag}\left(\lambda^r(\xi), \; \lambda^e(\xi) \right) \matr E(\xi)^T \vect v^N \\
  & =\left[\vect r(\xi) \; \vect e (\xi) \right] \text{diag}\left(\lambda^r(\xi), \; 0 \right) \left[\vect r(\xi) \; \vect e (\xi) \right]^T \vect v^N  \\
  & = \lambda^r(\xi) \left[\vect r(\xi) \; \vect 0 \right] \text{diag}(1, \; 0) \left[\vect r(\xi) \; \vect 0 \right]^T \vect v^N \\
  & = \lambda^r(\xi) \left( \vect r(\xi)^T \cdot  \vect v^N \right)  \vect r(\xi)
\end{align*}
Furthermore, from (\ref{eq:eigenvalues_simples}) we have
\begin{equation}
  \lambda^r(\xi, \vect v^N)
  \begin{cases}
    > 0 & \text{if} \; \dotprod{\vect r(\xi)}{\vect v^N}  > 0 \\
    \leq 0 & \text{otherwise}
  \end{cases} \label{eq:lambda_r_limits}
\end{equation}
Consequently, the Neuman boundary condition holds true:
\begin{align*}
  \vect n(\xi)^T  \, \dot{\xi} &= \vect n(\xi)^T  \, \left( \lambda^r) \dotprod{\vect r(\xi)}{\vect v^N} \vect r(\xi) \right)  \\
  & \geq 
  \underbrace{\text{arccos}\left(\delta / 2\right)}_{\approx 1, \; \text{since} \; \delta \ll 1} \underbrace{
  \lambda^r \dotprod{\vect r(\xi)}{\vect v^N}}_{\geq 0, \;\text{from} \; (\ref{eq:lambda_r_limits})} \geq 0
\end{align*}
As a result, we have a smooth vector field with no local minima away from the obstacles and proven impenetrability. 
${ }$ \hfill $\blacksquare$

\subsection{Practical Considerations} \label{sec:practical_considrations}
\subsubsection{Small Obstacles}
The discrete sampling of a laser scan might miss the closest point of an obstacle, i.e., an edge of an obstacle (see Fig.~\ref{fig:small_obstacles}). A small margin can be added to the obstacle to account for this:
\begin{equation*}
   R^{\mathrm{marg}} / R^{\mathrm{robo}} = \sin(\delta / 2) / \tan( \phi^{\min} / 2 ) + (1 - \cos(\delta / 2) )
\end{equation*}
Note that the first summand comes from the obstacle's corner and the second summand is the contribution of the agent's curvature. A good choice for the minimum obstacle angles is $\phi^{\min} =$ \SI{45}{\degree}. \\
Additionally, any obstacle with its robot to obstacle curvature radius ratio of
$R^{\mathrm{obs}} / R^{\mathrm{robo}} > \sin (\delta / 2) / \cos (\phi^{\mathrm{min}} / 2)$
is avoided safely, too (see Fig.~\ref{fig:small_obstacles}).

\subsubsection{Concave Regions}
When approaching a concave region, as in Fig.~\ref{fig:concave_region}, the algorithm is not aware that the resulting normal is obtained from multiple distinct walls. \\
Nevertheless, under the assumption that the distance to both walls is close to zero and that we obtain approximately the same weight from both walls, the resulting reference direction $\hat{\vect r}(\xi)$ points opposite the corner and has a large magnitude. It follows from (\ref{eq:eigenvalues_simples}), that the modulated DS will be pointing along the reference direction, i.e. $ \dot{\xi}^T \cdot \hat{\vect r}(\xi) / (\| \dot{\xi} \| \, \| \hat{\vect r}(\xi) \|) \rightarrow 1$. Hence we have to ensure collision avoidance.

\renewcommand*{\bibfont}{\small}
\printbibliography

\end{document}